\documentclass[letterpaper]{article}
\usepackage{aaai}
\usepackage{times}
\usepackage{helvet}
\usepackage{courier}
\usepackage{graphicx}
\usepackage{times}
\usepackage{latexsym}
\usepackage{booktabs}
\usepackage{amsmath}
\usepackage{amssymb}
\usepackage{graphicx}
\usepackage{bm}
\usepackage{CJKutf8}
\usepackage{amsmath}
\usepackage{url}
\usepackage{tabularx}
\usepackage{subfig}
\usepackage{multirow}

\usepackage{color}
\usepackage[ruled,linesnumbered]{algorithm2e}
\begin{document}
%
\title{Industry Risk Assessment via Hierarchical Financial Data Using Stock Market Sentiment Indicators}
\author{
Hongyin Zhu\\
hongyin\_zhu@163.com\\
}
\maketitle
\begin{CJK*}{UTF8}{gbsn}
\begin{abstract}
Risk assessment across industries is paramount for ensuring a robust and sustainable economy. While previous studies have relied heavily on official statistics for their accuracy, they often lag behind real-time developments. Addressing this gap, our research endeavors to integrate market microstructure theory with AI technologies to refine industry risk predictions. This paper presents an approach to analyzing industry trends leveraging real-time stock market data and generative small language models (SLMs). By enhancing the timeliness of risk assessments and delving into the influence of non-traditional factors such as market sentiment and investor behavior, we strive to develop a more holistic and dynamic risk assessment model. One of the key challenges lies in the inherent noise in raw data, which can compromise the precision of statistical analyses. Moreover, textual data about industry analysis necessitates a deeper understanding facilitated by pre-trained language models. To tackle these issues, we propose a dual-pronged approach to industry trend analysis: explicit and implicit analysis. For explicit analysis, we employ a hierarchical data analysis methodology that spans the industry and individual listed company levels. This strategic breakdown helps mitigate the impact of data noise, ensuring a more accurate portrayal of industry dynamics. In parallel, we introduce implicit analysis, where we pre-train an SML to interpret industry trends within the context of current news events. This approach leverages the extensive knowledge embedded in the pre-training corpus, enabling a nuanced understanding of industry trends and their underlying drivers. Experimental results based on our proposed methodology demonstrate its effectiveness in delivering robust industry trend analyses, underscoring its potential to revolutionize risk assessment practices across industries.
\end{abstract}

\section{Introduction}
Analyzing industry trends holds paramount importance for societal and economic advancement. The stock market serves as a vital indicator of the real economy's health, and thus, delving into its data offers profound insights for forecasting industry trends in the tangible world. By meticulously examining the sentiment within the capital market, we can gain a nuanced understanding of various industries' potential performance and future growth prospects. In the realm of behavioral finance, investor sentiment is intricately intertwined with the risk landscape of the respective industries. The sheer volume of transactional data in the stock market, coupled with the extensive textual data about economic developments, constitutes a multimodal data blend that necessitates rigorous analysis utilizing sophisticated AI tools. This multimodal approach ensures a comprehensive and nuanced assessment of industry dynamics, enabling stakeholders to make informed decisions and navigate the ever-evolving economic landscape with greater precision.

Previous research has relied heavily on mathematical statistical models and official data for industry analysis. For instance, Park et al. \cite{park2021impact} employed regression analysis to explore the interplay between technological prowess and financial performance in the semiconductor sector. However, a notable limitation of such analytical methods is their potential lack of timeliness. Traditional crisis detection models, such as the Value at Risk model \cite{borodovsky2000professional}, can become unstable when confronted with non-normal loss distributions (e.g., fat-tail events) or dynamic portfolio shifts. Similarly, the Expected Shortfall model \cite{nadarajah2014estimation} grapples with issues like computational intensity and data scarcity. To mitigate these challenges, this paper introduces stock market sentiment analysis as a novel risk detection indicator. We aim to leverage stock market data for industry trend analysis and prediction, thereby enhancing timeliness and improving accuracy. The question arises: how can investor sentiment in behavioral finance be accurately quantified using AI algorithms and seamlessly integrated into risk assessment models? To address this, we propose a dual-pronged approach to industry trend analysis: explicit analysis rooted in statistical methods and implicit analysis leveraging small language models. This paper presents an updated industry trend analysis framework, building upon the foundations laid in \cite{zhu2022financial}. Our framework is dedicated to thoroughly examining the development prospects, opportunities, and risks inherent in diverse industries. It incorporates advancements in statistical machine learning models and small language models for financial data analysis, offering a more nuanced and comprehensive perspective on industry dynamics.

Analyzing and forecasting industry trends through stock market data poses a multifaceted challenge, primarily centered around balancing data timeliness with stability. In the realm of explicit analysis grounded in statistical data, data of varying granularities exhibit distinct characteristics. Specifically, stock prices of listed companies or industry indexes are inherently time-sensitive, reacting swiftly to corporate performance fluctuations and news interventions, often resulting in heightened noise levels. Conversely, industry trend factors crafted using expert knowledge tend to be more stable but may lack the necessary sensitivity to capture timely changes. To reconcile these issues, we introduce a hierarchical data analysis approach. This method meticulously examines two levels of information: the intricate stock transaction data of listed companies and the industry trend factors designed by experts. By treating these two as complementary pieces of evidence, we aim to mitigate the influence of noise while preserving the essence of both data sources. This integrated analysis framework offers a more nuanced and robust perspective on industry trends, enhancing the accuracy and timeliness of our forecasts.

The second challenge we face is effectively pre-training a generative small language model (SLM) \cite{zhu2023metaaid,abdin2024phi} with textual data pertinent to industry analysis while ensuring its applicability to the ever-evolving context of current events and world news. Leveraging the advantages of SLMs, including low resource requirements, swift inference speeds, and flexibility, we embark on a pre-training process utilizing text data closely tied to industry trends, such as comprehensive industry research reports. This enriched corpus is used to further refine a GPT-2 model \cite{radford2019language,zhu2021collaborative}, enhancing its domain-specific knowledge \cite{zhu2024node}. To further tap into the industry insights captured by the model, we employ prompt learning techniques \cite{liu2023pre}, allowing us to probe and extract valuable industry knowledge in a more targeted manner. However, recognizing that our training corpus is limited to data before April 2022, we face a limitation in the model's ability to perceive and analyze current events, resulting in a potential disconnect from real-world developments. To address this gap, we incorporate retrieval augmented generation (RAG) \cite{gao2023retrieval} during the inference stage. By harnessing the power of the Google search engine, specifically utilizing the Serper API, we retrieve pertinent passages on current news events to enrich the input context of our model. This augmentation enables our model to minimize hallucinations and deliver more precise, up-to-date insights. By seamlessly fusing real-time context with the domain-specific knowledge of our SLM, we ensure that our analysis remains not only highly relevant but also nimble in adapting to the ever-evolving landscape of industry trends and global events.

Drawing upon this risk assessment framework, we developed a dedicated website that enables comprehensive analysis of China's A-share market data. This platform facilitates the examination of industry trends and risk assessments, contributing to the field in several aspects:

(1) We introduce a novel risk assessment framework that broadens the application of investor sentiment within behavioral finance and market microstructure theory. Our approach combines explicit analysis based on stock trading data with implicit analysis leveraging SLMs, thereby enabling precise industry trend analysis and forecasting.

(2) We present a hierarchical data analysis methodology that navigates the delicate balance between data stability and timeliness, ensuring that our conclusions are both robust and responsive to market dynamics.

(3) Through empirical analysis, we demonstrate the effectiveness of our framework by analyzing trends within specific industries and achieving commendable results. These achievements underscore the practical value and robustness of our proposed approach.

\section{Related Work}

\subsection{Industry Risk Assessment}
An et al. \cite{an2020russian} explored potential cooperation between South Africa and Russia in the energy sector, especially in energy exports and green energy. The study compared the advantages and disadvantages of coal, nuclear energy, and other renewable energy sources, and discussed opportunities for cooperation between the two countries in the energy sector. The study also looked at the potential of bioenergy and how it could contribute to the growth of energy demand through new technologies and population growth, stressed the importance of diversifying energy imports, and pointed out that energy cooperation is a key area of foreign policy and economic activity for both South African and Russian governments and private companies. They declared that strengthening energy cooperation between South Africa and Russia is very important. An et al. \cite{an2020strategy} explored South Korea's strategy in the global oil market, especially the potential for oil cooperation with Russia and the Democratic People's Republic of Korea (DPRK). The study used game theory to analyze this cooperation and believed that although oil cooperation is like a zero-sum game, both sides have common interests and can achieve a win-win situation. South Korea is committed to reducing its dependence on Middle Eastern oil and seeking to diversify its import sources, and Russia may become its main oil supplier. The study also mentioned that the new government in South Korea may have an impact on its cooperation with Russia and the DPRK. 

Candila et al. \cite{candila2021relationship} They explore the interdependence between exchange rates and oil earnings during the Great Recession and COVID-19 pandemic for 12 countries and regions that account for more than 60\% of global oil exports and 67\% of global imports. The study used the DCC-MIDAS model to analyze long-term correlation and found that the long-term correlation of oil exporting countries and the correlation of WTI crude oil returns were significantly enhanced after the epidemic. The correlation patterns vary across countries, with some countries turning positive and others remaining negative. A similar pattern is observed among oil-importing countries. Research reveals that there is a long-term correlation between oil prices and exchange rates, and oil price fluctuations affect exchange rates, which is of great significance for understanding the interaction between oil markets and financial markets and formulating macroeconomic policies. Saqib et al. \cite{saqib2021responses} studied the impact of exchange rate fluctuations on Pakistan's energy imports, especially crude oil, petroleum products, coal and electricity. The analysis using the NARDL model showed that exchange rate depreciation increases imports of crude oil and electricity in the short term and increases the import cost of all energy in the long term, but has the opposite effect on coal and petroleum products. The study recommends strengthening energy policies, improving domestic supply capacity, and stabilizing the exchange rate. Alwaely et al. \cite{alwaely2021emotional} studied the relationship between emotional knowledge and social skills in 4-5-year-old children. The subjects were 300 Sudanese kindergarten children, measured using the Emotion Matching Task and the Simplified Social Skills Assessment Scale. The results showed that emotional knowledge was positively correlated with social skills, and age and gender affected social development. Single-parent families and boys were more likely to have social and emotional problems.

Wen et al. \cite{wen2019stock} introduce a method to predict the trend of financial time series, using convolutional neural networks (CNN) to model the spatial correlation of historical trends and current trends, combined with frequent pattern mining. Ananthi et al.\cite{ananthi2021retracted} use k-NN regression for market trend prediction and came up with a set of technical indicators that the system used to predict the stock prices of many companies in the next few days. They collected predicted stock prices on specific stocks from a variety of sources and allowed the system to predict the overall sentiment of the stock. 
Nabipour et al. \cite{nabipour2020predicting} use machine learning and deep learning algorithms to predict stock market trends. They selected four stock market groups from the Tehran Stock Exchange: diversified finance, oil, non-metallic minerals, and basic metals, and took ten technical indicators from ten-year historical data as the input value. Each of the forecasting models will be evaluated by three indicators according to the input method, and the indicators will be calculated as stock trading volume and then converted into binary data.
Gandhmal et al. \cite{gandhmal2019systematic} review different techniques for stock market forecasting, and summarize them into two categories: forecasting techniques and clustering techniques, including Bayesian models, fuzzy classifiers, artificial neural networks, support vector machine classifiers, neural networks, fuzzy-based technology, and other methods. They analyze and elaborate on research gaps and issues in forecasting the stock market. 

Suryono et al. \cite{suryono2020challenges} provide an overview of Fintech research and categorize it by business process. They analyze the latest advances, potential gaps, trends, and challenges in fintech research. Chen \cite{chen2021stock} proposes a method for predicting short-term stock movements based on financial news headlines. The whole process includes news annotation, generation of PLM representation, training of neural network model, model verification of various indicators, and model output to guide trading strategies. They designed an RNN model that uses representations generated after fine-tuning BERT on domain corpora. They also introduced new indicators for assessing market news movements.
Mikhaylov \cite{mikhaylov2023understanding} examines the risks involved in wallets, depository services, trading, lending, and borrowing in the cryptocurrency space and offers recommendations for policymakers to implement a regulatory framework for digital assets that is comparable to the traditional financial system. Mutalimov et al. \cite{mutalimov2021assessing} study the development of small and micro enterprises in the Russian Far East, using mathematical models and regression analysis to understand indicators and regional differences. They analyze the impact of economic, social, and environmental sustainability, point out development challenges, recommend that the government formulate policies to support small and micro enterprises, and propose an evaluation method system. This research provides reference value and emphasizes that small and micro enterprises need comprehensive measures. 

\subsection{Financial Pre-trained Language Model}
Hiew et al. \cite{hiew2019bert} use BERT to construct a text-based financial sentiment index for three popular stocks listed on the Hong Kong exchange. We propose a financial sentiment analysis framework that maps textual sentiment to individual investor sentiment, option-implied sentiment to institutional investor opinion, and market-implied sentiment to the overall attitude of all market participants. 
Zhao et al. \cite{zhao2021bert} propose a BERT-based sentiment analysis and key entity detection method, which is applied to online financial text mining and social media public opinion analysis. They use a pre-trained model to perform sentiment analysis on web texts and then use labels as constraints to extract key entities and key entity detection is regarded as a sentence matching task. 
Zhang et al. \cite{zhang2022finbert} use machine reading comprehension methods for NER tasks in the financial domain. They introduce important prior information by utilizing well-crafted queries and extracting the start and end indexes of target entities. 

BloombergGPT \cite{wu2023bloomberggpt} contains 50 billion parameters. The model uses Bloomberg's extensive data sources to build a financial-specific dataset of 363 billion tokens and integrates a general dataset of 345 billion tokens. The model performs well in multiple benchmarks.
Li et al. \cite{li2023large} reviewed the existing application solutions of LLMs in the financial field, including zero-shot/few-shot learning, domain data fine-tuning, and training customized models from scratch. 
FinGPT \cite{yang2023fingpt} takes a data-centric approach, providing high-quality, transparent financial data resources, supporting researchers and practitioners to develop their own financial LLMs, and gaining support from the open-source AI4Finance community.
InvestLM \cite{yang2023investlm}, fine-tuned on LLaMA-65B using a carefully curated dataset of financial investment instructions, performs well in understanding financial texts and providing investment advice, and is highly evaluated by financial experts, comparable to commercial models.

\section{Approach}

Industry trend analysis focuses on unraveling the dynamics of industry trends ($Y$) by leveraging both industry-related textual data ($S$) and numerical data ($D$). For the numerical component, we predominantly employ mathematical modeling techniques to dissect $Y_1 = F_d(D)$, capturing the underlying relationships within the data. Meanwhile, for textual data, we harness SLMs to analyze $Y_2 = F_t(S)$, extracting insights from the implicit sentiments and nuances within the text. The cornerstone contribution of this paper lies in showcasing an approach to industry trend analysis that integrates both explicit analysis of statistical data and implicit analysis based on generative SLMs. We commence by delving into the explicit analysis method, elucidating how numerical data informs our understanding of industry trends. Subsequently, we introduce the implicit analysis method, illustrating how SLMs enrich our perspective by tapping into the rich sentiment and context embedded within textual data. This integrated approach offers a holistic view of industry trends, enhancing the accuracy and depth of our analysis.

\subsection{Explicit Analysis of Industry Trends}

This section introduces a hierarchical approach to industry trend analysis, which systematically examines trends at both the company and industry levels. By analyzing these distinct yet interconnected tiers, we gain a nuanced understanding of the dynamics shaping the overall industry landscape.

\subsubsection{Company Trend Analysis}
The stock price trend of a publicly traded company serves as a potent indicator, mirroring the company's developmental trajectory. In this subsection, we focus on conducting regression analysis to unravel the intricacies of a company's stock price dynamics. Predicting the movement of a company's stock price poses a formidable challenge, with most contemporary approaches centering on stock backtesting and short-term trend forecasting. The complexity of long-term stock price trend prediction stems from the scarcity of predictive factors that extend into the future. Drawing inspiration from the capabilities of recurrent neural networks \cite{Hochreiter1997LongSM} in handling sequential data, we introduce a recursive factor prediction method. This innovative approach recursively generates a long-term trend sequence, enabling us to capture the gradual evolution of stock price movements over an extended period. To validate the effectiveness of our method, we conduct experiments leveraging a regression model, aiming to accurately predict the long-term trend of a company's stock price, as formulated in Equation \ref{recursive.eq}.
\begin{align}
\label{recursive.eq}
x_i^{(t)} = f_i(x_1^{(t-1)}, x_2^{(t-1)},..., x_n^{(t-1)})
\end{align}
where $x_i$ represents the $i$-th factor. The superscript $^{(t)}$ represents the $t$-th time step and $x_i^{(0)}$ is $ x_i$. $f_i(\cdot)$ represents the prediction model for each factor.

Lasso regression, introduced by Tibshirani in 1996 \cite{tibshirani1996regression}, represents a refinement of linear regression through the integration of L1 regularization. This modification to the traditional linear regression objective function, as presented in Equation \ref{lasso.eq}, aims to minimize the sum of squared errors while simultaneously penalizing the absolute magnitude of the regression coefficients. This regularization technique promotes sparsity in the model, encouraging many coefficients to shrink toward zero, thus facilitating the identification of the most salient predictors and reducing the risk of overfitting.
\begin{align}
\label{lasso.eq}
L_i = \frac{1}{N}\sum_{j = 1}^n ( y_j - W_i^T X_j)^2 + \lambda \|W_i\|_1 
\end{align}
where $L_i$ represents the prediction model for the $i$-th factor, $W_i \in \mathbb{R}^d$ represents the corresponding weight vector, and $X_j\in \mathbb{R}^d$ represents the $j$-th training sample vector. $\lambda$ represents the coefficient of the regularization term.

One limitation of recursive factor prediction in the context of stock price trend analysis is the tendency for the predicted trend to converge towards a static value over time. To address this issue, we employ a hybrid approach that combines multiple models. This strategy involves iteratively predicting factors and recursively fitting the model, as previously outlined, to capture long-term trends more effectively. Furthermore, to enhance the realism of the predicted trend and mitigate excessive smoothness, we incorporate random factor disturbances into the model. By doing so, we aim to generate a fitting curve that more closely aligns with the intricate fluctuations observed in real-world stock price trends. This approach helps to preserve the dynamic nature of the market and provides a more nuanced view of potential future movements.
\begin{align}
x_i^{(t)} = x_i^{(t)} + \psi(\epsilon)* \alpha_i * x_i^{(t)} 
\end{align}
where the parameter $\alpha_i$ serves as a modulator, controlling the amplitude of fluctuations introduced into the predicted stock price trend. These fluctuations are sampled with a probability determined by $\epsilon$, which is drawn from a uniform distribution between 0 and 1. The function $\psi(\cdot)$, defined in Equation \ref{cases.eq}, acts as a piecewise mechanism, incorporating higher-order moments to regulate the probability of positive and negative deviations from the base trend. This nuanced approach acknowledges the emotional and liquidity-driven nuances of market dynamics. During times of heightened market sentiment, the likelihood of positive deviations increases, whereas periods of subdued emotions or reduced liquidity often coincide with negative deviations. By employing a threshold that is calibrated based on historical rise and fall statistics within the data, our method ensures that the probability and magnitude of sampling fluctuations align with realistic market behavior. In summary, our technique enhances the realism of curve fitting by dynamically adjusting the probability and scale of deviations, thereby producing a more nuanced and representative depiction of potential future stock price movements.
\begin{align}
\label{cases.eq}
\psi(x) = \begin{cases}1 & x > t_1 \\ 0 & t_2 \leq x \leq t_1 \\-1 & x < t_2 \end{cases}
\end{align}

Short-term predictions of stock prices are inherently vulnerable to disruptions arising from various external factors, including industry downturns, unforeseen geopolitical conflicts, and regulatory sanctions targeting specific sectors. These events often elicit abrupt and pronounced reactions in a company's share price, making short-term forecasting particularly challenging. However, from a long-term perspective, the overarching trend within industries tends to exhibit inherent cyclicality, necessitating a robust analysis that filters out noise to reveal underlying patterns. It is important to acknowledge the limitations of our method, which primarily relies on the capital market's response to such events and does not encapsulate the individual growth potential or prospects of listed companies. When delving into the specifics of a given listed company, a more comprehensive analysis that incorporates the company's financial performance data and other relevant metrics becomes imperative. Unfortunately, due to the constraints imposed by the length and scope of this paper, we are unable to delve into these nuances in depth. Nonetheless, we emphasize the significance of integrating such company-specific evaluations when formulating comprehensive long-term strategies.

\subsubsection{Industry Trend Analysis}
This subsection delves into industry-level analysis, which involves extracting industry-specific factors from the stock market to gain insights into broader industry trends. The approach offers a relatively stable framework for analysis, as it leverages expertly constructed evaluation indicators that capture the essence of industry dynamics. By focusing on these industry factors, we can gain a more holistic understanding of the trends and patterns shaping various sectors within the market.

\begin{figure}[!h]
\centering
\includegraphics[width=2.5in]{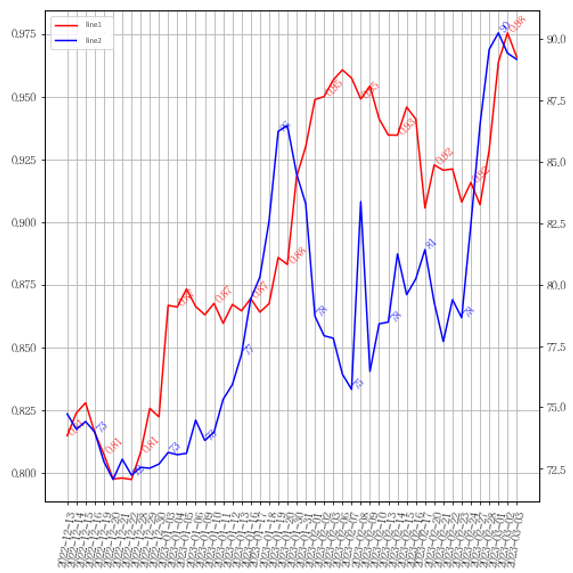}
\caption{Software industry trend analysis}
\label{industry1.fig}
\end{figure}

In our analysis, we have selected two distinct industries for comparison: software and coal. As depicted in Figure \ref{industry1.fig}, the red line illustrates the actual market trends for the software industry, showcasing a consistently positive and upward trajectory. Conversely, the blue line represents our projections for future industry trends, focusing on a conservative outlook over the next 3 to 6 months. It is evident from the chart that expectations for the software industry's future performance are on an upward trajectory, indicating optimism among market participants.

\begin{figure}[!h]
\centering
\includegraphics[width=2.5in]{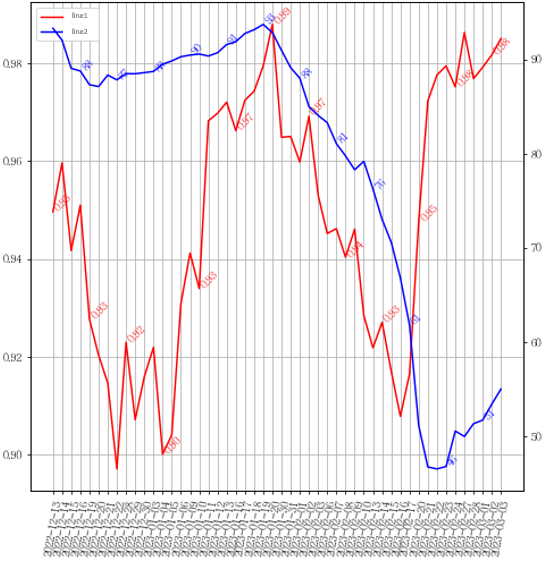}
\caption{Coal industry trend analysis}
\label{industry2.fig}
\end{figure}

As Figure \ref{industry2.fig} illustrates, the red line depicts a more volatile trend, underscoring the heightened uncertainty surrounding the current market trajectory of the coal industry. In contrast, the blue line indicates a gradual downward slope, suggesting that the coal industry is likely to face short-term pressures in its development. This disparity in trends highlights the differing outlooks for these two industries and underscores the importance of industry-specific analysis in navigating market dynamics.

\subsection{Implicit analysis of industry trends}
This subsection meticulously explores the refinement of a generative small language model, notably GPT-2, through a further pre-training phase \cite{zhu2023pre} utilizing unstructured textual data \cite{zhu2022switchnet} sourced from the vast expanse of the Internet. This rich dataset comprises approximately 1 billion tokens, encompassing industry research reports spanning diverse sectors within China, as well as insightful expert analysis and opinions. The core objective is to illustrate the methodology adopted for leveraging this approach to meticulously analyze industry trends. GPT-2, an autoregressive language model deeply rooted in the transformer decoder architecture, forms the solid foundation upon which our analysis is built. By leveraging its capabilities, we can represent a language model in the following manner:
\begin{align}
\label{language.eq}
p(x) = \prod_{i=1}^n p(x_i|x_{i-1},...,x_1)
\end{align}
where $x$ is the input sequence and $x_i$ represents the $i$-th word. GPT-2 incorporates the task format as input during zero-shot learning, enabling it to directly generate predictions for downstream tasks without requiring any model fine-tuning, as demonstrated in Equation \eqref{prompt.eq}. Similarly, the few-shot learning approach utilizes both the task format and a limited number of examples as input, guiding the model to make predictions for downstream tasks in a direct manner.
\begin{align}
\label{prompt.eq}
\hat{y}=\arg\max p(y|[x_1,x_2,...,x_n];T)
\end{align}
where $\hat{y}$ is the generated response and $T$ is the task form.

GPT-2 is recognized as a versatile multi-task learner, having been trained on vast amounts of text data. Its capability to directly engage in prompt learning for various downstream tasks, without the need for fine-tuning, is particularly noteworthy. By subjecting GPT-2 to training on a corpus rich in industry trend-related content, we can integrate relevant knowledge into the model. The primary advantage of utilizing GPT-2 for industry trend analysis lies in its ability to respond to a wide array of user queries with open-ended answers. However, it is important to acknowledge that the model's responses may lack concrete evidence and can be challenging to interpret. Nevertheless, GPT-2 serves as a valuable tool for generating innovative ideas and facilitating the construction of datasets in this domain.

Large Language Models (LLMs) \cite{zhu2023climate}, exemplified by chatGPT \cite{openai2022chatgpt}, are capable of producing more convincing and fluid responses in relevant tasks. However, these models occasionally struggle with accurately comprehending financial entities, as evidenced by instances where chatGPT misinterprets ``Everbright Securities" (光大证券) as ``GF Securities" (广发证券). LLMs' knowledge base is limited to the point in time at which they were trained, prohibiting them from directly analyzing current industry trends. Moreover, due to their vast number of parameters, frequent fine-tuning of these models presents significant challenges.

To mitigate the limitation of lacking current news event context within the model, we employ the retrieval-augmented generation paradigm \cite{zhu2023reranking,gao2023retrieval}, which interfaces with the Google search engine. By incorporating the latest news events as part of the input, we provide the model with essential background knowledge. This approach enables the model to generate more natural and targeted answers, thereby enhancing its responsiveness to real-world developments.




\section{Experiment}
This section commences with a comprehensive presentation of the experimental setup, providing a clear blueprint for understanding the methodology employed. Following this, it delves into the results of two distinct analyses: explicit industry analysis and implicit industry analysis, offering a nuanced understanding of the insights gained through the experimental process.
\subsection{Setup}
We have trained statistical models utilizing comprehensive transaction data sourced from China's A-share market spanning the years 2020 to 2023. To augment the capabilities of GPT-2, we conducted an additional pre-training phase specifically on Chinese textual data related to industry analysis, which was published before April 2022. These rigorous experiments were conducted on a high-performance computing environment, featuring an AMD Ryzen-9 5900X Processor @ 3.7GHz with 128GB of memory, alongside RTX 4080 GPUs, each equipped with 16GB of dedicated graphics memory.
\subsection{Explicit Industry Trend Analysis Results}
\begin{figure}[!h]
\centering
\includegraphics[width=3in]{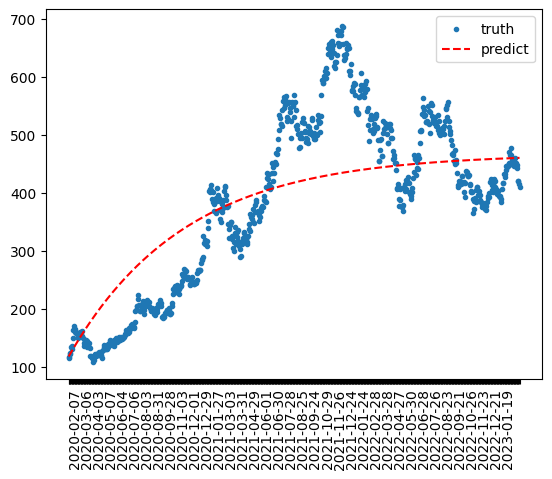}
\caption{Fitting and forecasting of stock price trend}
\label{ningde.fig}
\end{figure}

This subsection primarily showcases experiments conducted for explicit trend analysis, with a particular emphasis on company-level stock price forecasting. While industry-level trend analyses are elaborated upon in Figures \ref{industry1.fig} and \ref{industry2.fig}, the current experiment zeroes in on predicting stock price movements at the company level. To facilitate understanding, we utilize data from a representative listed company within a specific industry as a case study for conducting an industry trend analysis experiment. Figure \ref{ningde.fig} visually presents the outcomes of our proposed method, showcasing the recursively generated predictions. Here, "truth" denotes the actual stock price, whereas "predict" represents the stock price trend projected by our model. Notably, the method achieves an $R^2$ (coefficient of determination) score of 0.65, indicating a substantial degree of correlation between the predicted and observed trends.

\begin{figure}[!h]
\centering
\includegraphics[width=3in]{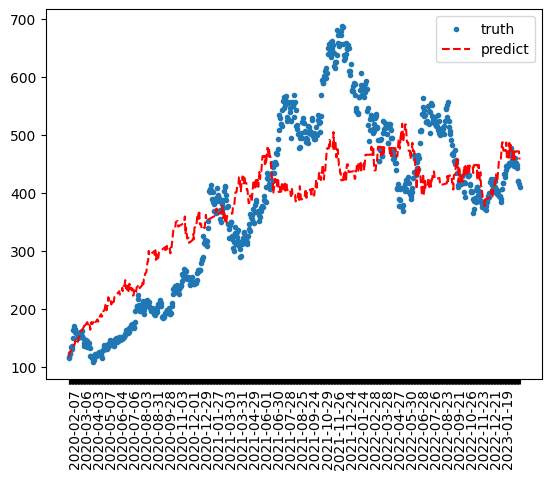}
\caption{Noisy stock price trend forecasting}
\label{ningderandom.fig}
\end{figure}

Figure \ref{ningderandom.fig} illuminates the impact of incorporating a noise strategy into our model. This modification results in a statistical $R^2$ score of 0.50, with a range of 0.15 across 10 random iterations. By introducing noise into the regression process, we aim to address the issue of the model potentially converging towards a static value, thereby enhancing its ability to more accurately capture the dynamic fluctuations inherent in real-world stock price trends.

\begin{figure}[!h]
\centering
\includegraphics[width=3in]{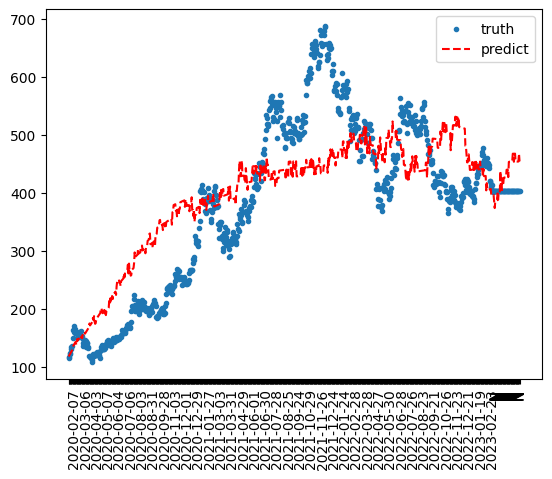}
\caption{Recursive factor prediction}
\label{ningdefuture.fig}
\end{figure}

In this analysis, our paper presents a forecast of stock price trends for a period extending 50 days beyond the dataset's coverage. The corresponding results are showcased in Figure \ref{ningdefuture.fig}, which demonstrates the efficacy of the proposed methods in generating insightful predictions. It is important to clarify that these findings are solely intended for academic discourse and should not be construed as investment advice or recommendations.

\subsection{Implicit Industry Trend Analysis Results}
This subsection delves into industry trend analysis experiments leveraging SLMs. Given the inherent variability and openness in the responses generated by SLMs, we employ a case study approach to decipher the underlying industry trends. Recognizing that the model may have limited awareness of real-world occurrences, we incorporate background knowledge by posing questions that are grounded in actual events that have transpired. By doing so, we prompt the model to generate responses that are relevant and informative, thereby facilitating a deeper understanding of the industry trends at play.

We have conducted comprehensive case studies utilizing three distinct versions of GPT-2: the Chinese variant (zh), the English version (en), and GPT-2 (industry) which underwent additional pre-training. To demonstrate their capabilities, we tasked these models with conducting trend analyses for the burgeoning industries of new energy and artificial intelligence. The responses generated by each model are summarized in Tables \ref{casestudy1} and \ref{casestudy2}. Upon examination, it becomes evident that GPT-2 (industry), owing to its specialized pre-training, provides answers that are not only more comprehensive but also exhibit a higher degree of logical coherence and depth compared to the other two models.

\begin{table*}[!ht]
\caption{Risk assessment of diverse models in the new energy industry}
\centering
\tiny
\resizebox{\linewidth}{!}{
    \begin{tabularx}{\linewidth}{p{1.0cm}|X}
    \toprule
    Settings   & Content \\ 
    \midrule
    Question & Foreign media commentary: China's strong economic rebound has enhanced global optimism, and analysts are optimistic about the future of the world's second largest economy. He said at the briefing: "We saw that retail sales began to pick up significantly in February. China is returning to growth. Not only online sales are recovering, but physical store sales are also recovering." What is the development prospect of the new energy industry? \\ \hline\hline
    GPT-2 (zh) & WSJ: We'll see. From the analysis of new energy vehicles, it can be seen that the entry of the automobile industry is the result of the promotion process of domestic new energy vehicles. The China Automobile Association stated that in the first half of this year, the sales of new energy vehicles in China reached 20,000, a year-on-year increase of 50\%. This speed is astonishing, and this magnitude is leading in the world. After the emergence of the new energy vehicle industry and its explosive growth, the total investment in new energy vehicles has increased significantly this year. China's auto industry has invested more than 20 billion yuan. In the field of new energy and new materials, the United States, Germany, Japan \\ \hline
    GPT-2 (en) & The new coal and petroleum industries are both expected to form the basis of a new type of China-Bolender coal production from the second half of the century, under heavy government pressure. The country is already having the first and second largest oil discoveries. The project's potential to provide new employment has been made possible by the development of new fields in offshore oil fields. China says it hopes to produce about 500,000 barrels oil a day by 2020 – about as much as the amount consumed today. The country's coal output has also increased markedly, with the first major coal mine built in 2005 being discovered recently. Beijing also has the final step in a major project to expand its capacity to over 20 megawatts by 2025. In its new energy policy, Beijing recently revised its policy on energy-related matters – more recently this month it introduced a range of policy directives to support the development of coal-fired power stations and an additional 14 energy-related energy services (EMS) to help fuel its efforts to shift away from coal-fired power stations. A China-Bolender Coal 
    \\\hline
    \textbf{GPT-2 (industry)} & We all know that batteries have a high cost, and now the cost of batteries has accounted for 20\% of the country's electricity consumption, but the country still has many bottlenecks in this technology, including the cost of charging piles. The level has improved, because now there is hydrogen energy, which can be called lithium batteries all over the world, so in this field, they have huge room for development. If we do not accelerate China's economic growth, the market prospect of our country will be It will be greatly discounted, so for new energy, how should we deal with it? First of all, BYD is in the field of motor vehicles  \\ 
    \bottomrule
    \end{tabularx}
}
\label{casestudy1}
\end{table*}

\begin{table*}[!ht]
\caption{Risk assessment of diverse models in the artificial intelligence Industry}
\centering
\tiny
\resizebox{\linewidth}{!}{
    \begin{tabularx}{\linewidth}{p{1.0cm}|X}
    \toprule
    Settings   & Content \\ 
    \midrule
    Question & The report notes that consumer demand, market conditions, and new artificial intelligence (AI) and machine learning (ML) technologies are all evolving, and increasing data complexity is creating a different picture than it was a year ago. According to the report, by 2026, 7PB of data will be generated every second globally. What is the future of artificial intelligence?  \\ \hline\hline
    GPT-2 (zh) & "Forbes" magazine rated it as the world's first research report on artificial intelligence in the field of artificial intelligence. The main purpose of the report is: the future trend of artificial intelligence is to use artificial intelligence to serve the future society of mankind. The report is not an isolated case, and its development of artificial intelligence has made an understanding of the future society. How much impact will the development of artificial intelligence have on the development of society? What kind of success can artificial intelligence ultimately achieve? What impact and impact will they have on all human beings? First of all, the development trend of artificial intelligence. The key to the success of artificial intelligence is based on big data, based on artificial intelligence, machine  \\ \hline
    GPT-2 (en) & The report continues: The report states, "[e]ven, today there is an increasing demand for artificial intelligence (AI) and ML technologies that enable users to search for, solve, and optimize complex tasks. This is the key to driving the development of intelligent devices and applications." In other words, new data challenges are occurring and demand for such applications is increasing worldwide. These trends, noted the report, could be "welcome news for consumers and companies and provide more opportunities for enterprises to utilize these technologies to improve consumer experience." In this context, artificial intelligence will be even more powerful in our economy as this will make the "digital economy" exponentially more diverse, and less connected. Google recently started using automated searches within their Gmail and Google Drive services as a way to save the company money. This means that Google Drive users will be able to browse all of Google's Google services, including their favorite videos, music, and pictures, without having to search for their entire inbox. As a result, the data it's capturing will be used as Google's way to "save money in terms of Google's online 
    \\\hline
    \textbf{GPT-2 (industry)} & At present, all scientific and technological innovations are driven by wisdom. They all develop towards data-driven development, and they can also develop towards networkization. Therefore, for the global market, everyone has seen the global prospect, but the global market prospect is still unclear. There is a long way to go, and the global intelligent system is also actively investing, but we still have to correctly understand the global economic policy, so we are here, we also deeply understand that economic policy is the top priority of policy, It is the leader of policy development. Global economic policy should play a decisive role in the allocation of global resources and promote common prosperity.  \\ 
    \bottomrule
    \end{tabularx}
}
\label{casestudy2}
\end{table*}


\section{Conclusion and future work}
This paper introduces an innovative risk assessment framework that expands the utilization of investor sentiment within the realms of behavioral finance and market microstructure theory. We present two complementary methods for analyzing industry trends based on stock market data: explicit analysis and implicit analysis. For explicit analysis, we employ regression models to recursively generate forecasts, ensuring a close alignment with current trends. For implicit analysis, we harness the power of generative SLMs to delve into the nuances of market sentiment, incorporating the context of current events for enhanced knowledge detection. The experimental outcomes demonstrate that while explicit analysis accurately mirrors prevailing trends, implicit analysis offers a more open-ended and insightful perspective, albeit with room for improvement in terms of controllability.

The key innovation of this work lies in its dual-pronged approach to industry trend analysis, marrying traditional data analysis with generative AI capabilities. We envision that future research will further bolster the capability of SLMs in analyzing industry trends, unlocking even deeper insights into the complexities of the stock market.
\end{CJK*}
\bibliographystyle{aaai}
\bibliography{reference}

\begin{thebibliography}{}

\bibitem[\protect\citeauthoryear{Abdin \bgroup et al\mbox.\egroup
  }{2024}]{abdin2024phi}
Abdin, M.; Jacobs, S.~A.; Awan, A.~A.; Aneja, J.; Awadallah, A.; Awadalla, H.;
  Bach, N.; Bahree, A.; Bakhtiari, A.; Behl, H.; et~al.
\newblock 2024.
\newblock Phi-3 technical report: A highly capable language model locally on
  your phone.
\newblock {\em arXiv preprint arXiv:2404.14219}.

\bibitem[\protect\citeauthoryear{Alwaely, Yousif, and
  Mikhaylov}{2021}]{alwaely2021emotional}
Alwaely, S.~A.; Yousif, N. B.~A.; and Mikhaylov, A.
\newblock 2021.
\newblock Emotional development in preschoolers and socialization.
\newblock {\em Early child development and care} 191(16):2484--2493.

\bibitem[\protect\citeauthoryear{An and Mikhaylov}{2020}]{an2020russian}
An, J., and Mikhaylov, A.
\newblock 2020.
\newblock Russian energy projects in south africa.
\newblock {\em Journal of Energy in Southern Africa} 31(3):58--64.

\bibitem[\protect\citeauthoryear{An, Mikhaylov, and
  Jung}{2020}]{an2020strategy}
An, J.; Mikhaylov, A.; and Jung, S.-U.
\newblock 2020.
\newblock The strategy of south korea in the global oil market.
\newblock {\em Energies} 13(10):2491.

\bibitem[\protect\citeauthoryear{Ananthi and
  Vijayakumar}{2021}]{ananthi2021retracted}
Ananthi, M., and Vijayakumar, K.
\newblock 2021.
\newblock Retracted article: stock market analysis using candlestick regression
  and market trend prediction (ckrm).
\newblock {\em Journal of Ambient Intelligence and Humanized Computing}
  12(5):4819--4826.

\bibitem[\protect\citeauthoryear{Borodovsky and
  Lore}{2000}]{borodovsky2000professional}
Borodovsky, L., and Lore, M.
\newblock 2000.
\newblock {\em Professional's Handbook of Financial Risk Management}.
\newblock Elsevier.

\bibitem[\protect\citeauthoryear{Candila \bgroup et al\mbox.\egroup
  }{2021}]{candila2021relationship}
Candila, V.; Maximov, D.; Mikhaylov, A.; Moiseev, N.; Senjyu, T.; and Tryndina,
  N.
\newblock 2021.
\newblock On the relationship between oil and exchange rates of oil-exporting
  and oil-importing countries: From the great recession period to the covid-19
  era.
\newblock {\em Energies} 14(23):8046.

\bibitem[\protect\citeauthoryear{Chen}{2021}]{chen2021stock}
Chen, Q.
\newblock 2021.
\newblock Stock movement prediction with financial news using contextualized
  embedding from bert.
\newblock {\em arXiv preprint arXiv:2107.08721}.

\bibitem[\protect\citeauthoryear{Gandhmal and
  Kumar}{2019}]{gandhmal2019systematic}
Gandhmal, D.~P., and Kumar, K.
\newblock 2019.
\newblock Systematic analysis and review of stock market prediction techniques.
\newblock {\em Computer Science Review} 34:100190.

\bibitem[\protect\citeauthoryear{Gao \bgroup et al\mbox.\egroup
  }{2023}]{gao2023retrieval}
Gao, Y.; Xiong, Y.; Gao, X.; Jia, K.; Pan, J.; Bi, Y.; Dai, Y.; Sun, J.; and
  Wang, H.
\newblock 2023.
\newblock Retrieval-augmented generation for large language models: A survey.
\newblock {\em arXiv preprint arXiv:2312.10997}.

\bibitem[\protect\citeauthoryear{Hiew \bgroup et al\mbox.\egroup
  }{2019}]{hiew2019bert}
Hiew, J. Z.~G.; Huang, X.; Mou, H.; Li, D.; Wu, Q.; and Xu, Y.
\newblock 2019.
\newblock Bert-based financial sentiment index and lstm-based stock return
  predictability.
\newblock {\em arXiv preprint arXiv:1906.09024}.

\bibitem[\protect\citeauthoryear{Hochreiter and
  Schmidhuber}{1997}]{Hochreiter1997LongSM}
Hochreiter, S., and Schmidhuber, J.
\newblock 1997.
\newblock Long short-term memory.
\newblock {\em Neural Computation} 9:1735--1780.

\bibitem[\protect\citeauthoryear{Li \bgroup et al\mbox.\egroup
  }{2023}]{li2023large}
Li, Y.; Wang, S.; Ding, H.; and Chen, H.
\newblock 2023.
\newblock Large language models in finance: A survey.
\newblock In {\em Proceedings of the fourth ACM international conference on AI
  in finance},  374--382.

\bibitem[\protect\citeauthoryear{Liu \bgroup et al\mbox.\egroup
  }{2023}]{liu2023pre}
Liu, P.; Yuan, W.; Fu, J.; Jiang, Z.; Hayashi, H.; and Neubig, G.
\newblock 2023.
\newblock Pre-train, prompt, and predict: A systematic survey of prompting
  methods in natural language processing.
\newblock {\em ACM Computing Surveys} 55(9):1--35.

\bibitem[\protect\citeauthoryear{Mikhaylov}{2023}]{mikhaylov2023understanding}
Mikhaylov, A.
\newblock 2023.
\newblock Understanding the risks associated with wallets, depository services,
  trading, lending, and borrowing in the crypto space.
\newblock {\em Journal of Infrastructure, Policy and Development} 7(3).

\bibitem[\protect\citeauthoryear{Mutalimov \bgroup et al\mbox.\egroup
  }{2021}]{mutalimov2021assessing}
Mutalimov, V.; Kovaleva, I.; Mikhaylov, A.; and Stepanova, D.
\newblock 2021.
\newblock Assessing regional growth of small business in russia.
\newblock {\em Entrepreneurial Business and Economics Review} 9(3):119--133.

\bibitem[\protect\citeauthoryear{Nabipour \bgroup et al\mbox.\egroup
  }{2020}]{nabipour2020predicting}
Nabipour, M.; Nayyeri, P.; Jabani, H.; Shahab, S.; and Mosavi, A.
\newblock 2020.
\newblock Predicting stock market trends using machine learning and deep
  learning algorithms via continuous and binary data; a comparative analysis.
\newblock {\em IEEE Access} 8:150199--150212.

\bibitem[\protect\citeauthoryear{Nadarajah, Zhang, and
  Chan}{2014}]{nadarajah2014estimation}
Nadarajah, S.; Zhang, B.; and Chan, S.
\newblock 2014.
\newblock Estimation methods for expected shortfall.
\newblock {\em Quantitative Finance} 14(2):271--291.

\bibitem[\protect\citeauthoryear{OpenAI}{2022}]{openai2022chatgpt}
OpenAI, T.
\newblock 2022.
\newblock Chatgpt: Optimizing language models for dialogue.
\newblock {\em OpenAI}.

\bibitem[\protect\citeauthoryear{Park \bgroup et al\mbox.\egroup
  }{2021}]{park2021impact}
Park, J.~H.; Chung, H.; Kim, K.~H.; Kim, J.~J.; and Lee, C.
\newblock 2021.
\newblock The impact of technological capability on financial performance in
  the semiconductor industry.
\newblock {\em Sustainability} 13(2):489.

\bibitem[\protect\citeauthoryear{Radford \bgroup et al\mbox.\egroup
  }{2019}]{radford2019language}
Radford, A.; Wu, J.; Child, R.; Luan, D.; Amodei, D.; and Sutskever, I.
\newblock 2019.
\newblock Language models are unsupervised multitask learners.

\bibitem[\protect\citeauthoryear{Saqib \bgroup et al\mbox.\egroup
  }{2021}]{saqib2021responses}
Saqib, A.; Chan, T.-H.; Mikhaylov, A.; and Lean, H.~H.
\newblock 2021.
\newblock Are the responses of sectoral energy imports asymmetric to exchange
  rate volatilities in pakistan? evidence from recent foreign exchange regime.
\newblock {\em Frontiers in Energy Research} 9:614463.

\bibitem[\protect\citeauthoryear{Suryono, Budi, and
  Purwandari}{2020}]{suryono2020challenges}
Suryono, R.~R.; Budi, I.; and Purwandari, B.
\newblock 2020.
\newblock Challenges and trends of financial technology (fintech): a systematic
  literature review.
\newblock {\em Information} 11(12):590.

\bibitem[\protect\citeauthoryear{Tibshirani}{1996}]{tibshirani1996regression}
Tibshirani, R.
\newblock 1996.
\newblock Regression shrinkage and selection via the lasso.
\newblock {\em Journal of the Royal Statistical Society: Series B
  (Methodological)} 58(1):267--288.

\bibitem[\protect\citeauthoryear{Wen \bgroup et al\mbox.\egroup
  }{2019}]{wen2019stock}
Wen, M.; Li, P.; Zhang, L.; and Chen, Y.
\newblock 2019.
\newblock Stock market trend prediction using high-order information of time
  series.
\newblock {\em Ieee Access} 7:28299--28308.

\bibitem[\protect\citeauthoryear{Wu \bgroup et al\mbox.\egroup
  }{2023}]{wu2023bloomberggpt}
Wu, S.; Irsoy, O.; Lu, S.; Dabravolski, V.; Dredze, M.; Gehrmann, S.; Kambadur,
  P.; Rosenberg, D.; and Mann, G.
\newblock 2023.
\newblock Bloomberggpt: A large language model for finance.
\newblock {\em arXiv preprint arXiv:2303.17564}.

\bibitem[\protect\citeauthoryear{Yang, Liu, and Wang}{2023}]{yang2023fingpt}
Yang, H.; Liu, X.-Y.; and Wang, C.~D.
\newblock 2023.
\newblock Fingpt: Open-source financial large language models.
\newblock {\em arXiv preprint arXiv:2306.06031}.

\bibitem[\protect\citeauthoryear{Yang, Tang, and Tam}{2023}]{yang2023investlm}
Yang, Y.; Tang, Y.; and Tam, K.~Y.
\newblock 2023.
\newblock Investlm: A large language model for investment using financial
  domain instruction tuning.
\newblock {\em arXiv preprint arXiv:2309.13064}.

\bibitem[\protect\citeauthoryear{Zhang and Zhang}{2022}]{zhang2022finbert}
Zhang, Y., and Zhang, H.
\newblock 2022.
\newblock Finbert-mrc: financial named entity recognition using bert under the
  machine reading comprehension paradigm.
\newblock {\em arXiv preprint arXiv:2205.15485}.

\bibitem[\protect\citeauthoryear{Zhao \bgroup et al\mbox.\egroup
  }{2021}]{zhao2021bert}
Zhao, L.; Li, L.; Zheng, X.; and Zhang, J.
\newblock 2021.
\newblock A bert based sentiment analysis and key entity detection approach for
  online financial texts.
\newblock In {\em 2021 IEEE 24th International Conference on Computer Supported
  Cooperative Work in Design (CSCWD)},  1233--1238.
\newblock IEEE.

\bibitem[\protect\citeauthoryear{Zhu and Tiwari}{2023}]{zhu2023climate}
Zhu, H., and Tiwari, P.
\newblock 2023.
\newblock Climate change from large language models.
\newblock {\em arXiv preprint arXiv:2312.11985}.

\bibitem[\protect\citeauthoryear{Zhu \bgroup et al\mbox.\egroup
  }{2021}]{zhu2021collaborative}
Zhu, H.; Tiwari, P.; Ghoneim, A.; and Hossain, M.~S.
\newblock 2021.
\newblock A collaborative ai-enabled pretrained language model for aiot domain
  question answering.
\newblock {\em IEEE Transactions on Industrial Informatics} 18(5):3387--3396.

\bibitem[\protect\citeauthoryear{Zhu \bgroup et al\mbox.\egroup
  }{2022}]{zhu2022switchnet}
Zhu, H.; Tiwari, P.; Zhang, Y.; Gupta, D.; Alharbi, M.; Nguyen, T.~G.; and
  Dehdashti, S.
\newblock 2022.
\newblock Switchnet: A modular neural network for adaptive relation extraction.
\newblock {\em Computers and Electrical Engineering} 104:108445.

\bibitem[\protect\citeauthoryear{Zhu \bgroup et al\mbox.\egroup
  }{2023}]{zhu2023pre}
Zhu, H.; Peng, H.; Lyu, Z.; Hou, L.; Li, J.; and Xiao, J.
\newblock 2023.
\newblock Pre-training language model incorporating domain-specific
  heterogeneous knowledge into a unified representation.
\newblock {\em Expert Systems with Applications} 215:119369.

\bibitem[\protect\citeauthoryear{Zhu}{2022}]{zhu2022financial}
Zhu, H.
\newblock 2022.
\newblock Financial data analysis application via multi-strategy text
  processing.
\newblock {\em arXiv preprint arXiv:2204.11394}.

\bibitem[\protect\citeauthoryear{Zhu}{2023a}]{zhu2023metaaid}
Zhu, H.
\newblock 2023a.
\newblock Metaaid 2.5: A secure framework for developing metaverse applications
  via large language models.
\newblock {\em arXiv preprint arXiv:2312.14480}.

\bibitem[\protect\citeauthoryear{Zhu}{2023b}]{zhu2023reranking}
Zhu, H.
\newblock 2023b.
\newblock Reranking passages with coarse-to-fine neural retriever using
  list-context information.
\newblock {\em arXiv preprint arXiv:2308.12022}.

\bibitem[\protect\citeauthoryear{Zhu}{2024}]{zhu2024node}
Zhu, H.
\newblock 2024.
\newblock Node classification via semantic-structural attention-enhanced graph
  convolutional networks.
\newblock {\em arXiv preprint arXiv:2403.16033}.

\end{thebibliography}

\end{document}